\documentclass[letterpaper, 10 pt, conference]{ieeeconf}  % Comment this line out if you need a4paper

\IEEEoverridecommandlockouts                              % This command is only needed if 
\usepackage{graphicx}
\usepackage{amsmath, amssymb}
\usepackage{makecell}
\usepackage{multirow}
\usepackage{subfigure}
\usepackage{url}
\usepackage{threeparttable}
\usepackage{hyperref}
\usepackage{cite}
\usepackage{bm}
\usepackage{pdfpages}
\usepackage{mathrsfs}
\newcommand{\zstroke}{\text{\ooalign{\hidewidth -\kern-.3em-\hidewidth\cr$z$\cr}}}
\allowdisplaybreaks

\DeclareGraphicsExtensions{.png,.pdf}

\title{\LARGE \bf
Haptic-ACT: Bridging Human Intuition with Compliant Robotic Manipulation via Immersive VR
}

\author{Kelin Li, Shubham M Wagh, Nitish Sharma, Saksham Bhadani, Wei Chen, \\Chang Liu, and Petar Kormushev% <-this % stops a space
\thanks{Kelin Li is jointly with the Robot Intelligence Lab, Imperial College London, and the Extend Robotics, {\tt\small k.li20@imperial.ac.uk}. Wei Chen, and Petar Kormushev are with the Robot Intelligence Lab, Imperial College London, 25 Exhibition Road, London, SW7 2DB, UK, {\tt\small (w.chen21, p.kormushev)@imperial.ac.uk}. Shubham M Wagh, Nitish Sharma, Saksham Bhadani, and Chang Liu are with Extend Robotics, 5-9 Merchants Pl, Reading, RG1 1DT, UK {\tt\small (shubham.wagh, nitish, saksham.bhadani, chang.liu)@extendrobotics.com}.}
}

\begin{document}

\maketitle
\thispagestyle{empty}
\pagestyle{empty}

%%%%%%%%%%%%%%%%%%%%%%%%%%%%%%%%%%%%%%%%%%%%%%%%%%%%%%%%%%%%%%%%%%%%%%%%%%%%%%%%
\begin{abstract}
Robotic manipulation is essential for the widespread adoption of robots in industrial and home settings and has long been a focus within the robotics community. Advances in artificial intelligence have introduced promising learning-based methods to address this challenge, with imitation learning emerging as particularly effective. However, efficiently acquiring high-quality demonstrations remains a challenge. In this work, we introduce an immersive VR-based teleoperation setup designed to collect demonstrations from a remote human user. We also propose an imitation learning framework called Haptic Action Chunking with Transformers (Haptic-ACT). To evaluate the platform, we conducted a pick-and-place task and collected 50 demonstration episodes. Results indicate that the immersive VR platform significantly reduces demonstrator fingertip forces compared to systems without haptic feedback, enabling more delicate manipulation. Additionally, evaluations of the Haptic-ACT framework in both the MuJoCo simulator and on a real robot demonstrate its effectiveness in teaching robots more compliant manipulation compared to the original ACT. Additional materials are available at \href{https://sites.google.com/view/hapticact}{https://sites.google.com/view/hapticact}.
\end{abstract}

%%%%%%%%%%%%%%%%%%%%%%%%%%%%%%%%%%%%%%%%%%%%%%%%%%%%%%%%%%%%%%%%%%%%%%%%%%%%%%%%
\section{Introduction}\label{sec:introduction}
% - 1. The topic of this research
% - 2. Current work and drawbacks
% - 3. What is our work
% - 4. The benefit of our work
With the growing demand for robotics to assist humans in daily manipulation tasks, robotic manipulation has garnered increasing attention from the robotics community. Over the past decades, it has made tremendous progress~\cite{Cui2021TowardManipulation,Shridhar2021CLIPORT:Manipulation,Johns2021Coarse-to-FineDemonstration,Li2022EfficientGrasp:Hands,Wu2022LearningVisualization,Kwon2024LanguageGenerators}. In these studies, robotic manipulation is typically performed by a robot arm equipped with a gripper attached to its end-effector. RGB-D cameras are commonly used to observe the environment and capture visual information, including the poses and geometric features of objects. These features can be represented as either 2D RGB images~\cite{Jiang2023VIMA:Prompts,Yu2022SE-ResUNet:Method,An2024RGBManip:Estimation} or 3D point clouds~\cite{Ni2020PointNet++Clouds,Liang2019PointNetGPD:Sets}, valid manipulations will be generated based on the observed object features. With the development of learning-based methods, efficiency and generalizability have become important considerations in the design of frameworks~\cite{Zhao2021REGNet:Clouds,Shao2020UniGrasp:Hands,Li2022EfficientGrasp:Hands}. In recent years, language models have been integrated with visual models to enable robots to handle a wide range of environments~\cite{Jiang2023VIMA:Prompts,Kwon2024LanguageGenerators,Duan2024Manipulate-Anything:Models}.

\begin{figure}[t!]
    \centering
    \includegraphics[width=\columnwidth]{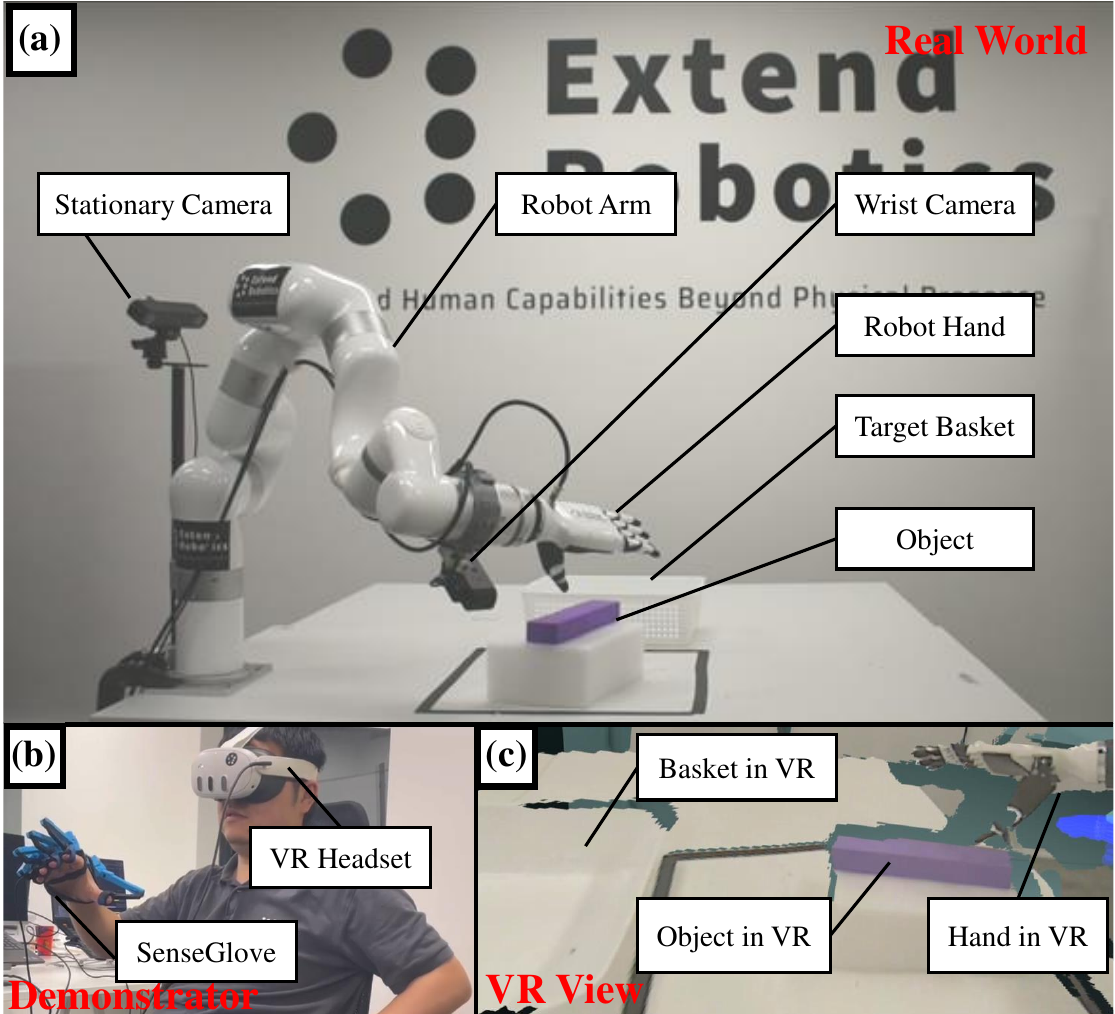}
    \caption{Summary diagram of the proposed immersive VR-based setup used in this work, featuring a VR headset, a haptic feedback glove, a follower robot arm, and a robot hand. (a) illustrates the robot arm and hand system following human demonstrations and providing sensory feedback, (b) depicts the demonstrator remotely controlling the robot, and (c) displays the VR view from the headset.}
    \label{fig:1}
    \vspace{-4mm}
\end{figure}

Although existing robotic manipulation methods can produce stable actions, a gap remains in applying traditional robot learning methods to real-world setups. Traditional robot learning methods involve the robot exploring the entire manipulation space to find a solution for a specific task. However, this process is usually inefficient and time-consuming, as it often involves redundant learning before arriving at an optimal solution. An efficient alternative to training robots in manipulation tasks is imitation learning, also known as Learning from Demonstration (LfD). In this approach, the robot learns by observing expert demonstrations, allowing skills to generalize to unseen scenarios. This process not only extracts information about the expert's behavior and the environment but also learns the mapping between observations and actions.~\cite{Hua2021LearningLearning}. Thus, the robot can learn in the correct direction to perform manipulation tasks effectively. In recent years, imitation learning has been extensively studied for enabling robots to perform various manipulation tasks~\cite{Kim2021Transformer-basedManipulation,Belkhale2023HYDRA:Learning,Xie2024DecomposingManipulation}. However, efficiently collecting demonstrations using an appropriate platform remains a challenge.

\begin{figure*}
    \vspace{4mm}
    \centering
    \includegraphics[width=0.9\textwidth]{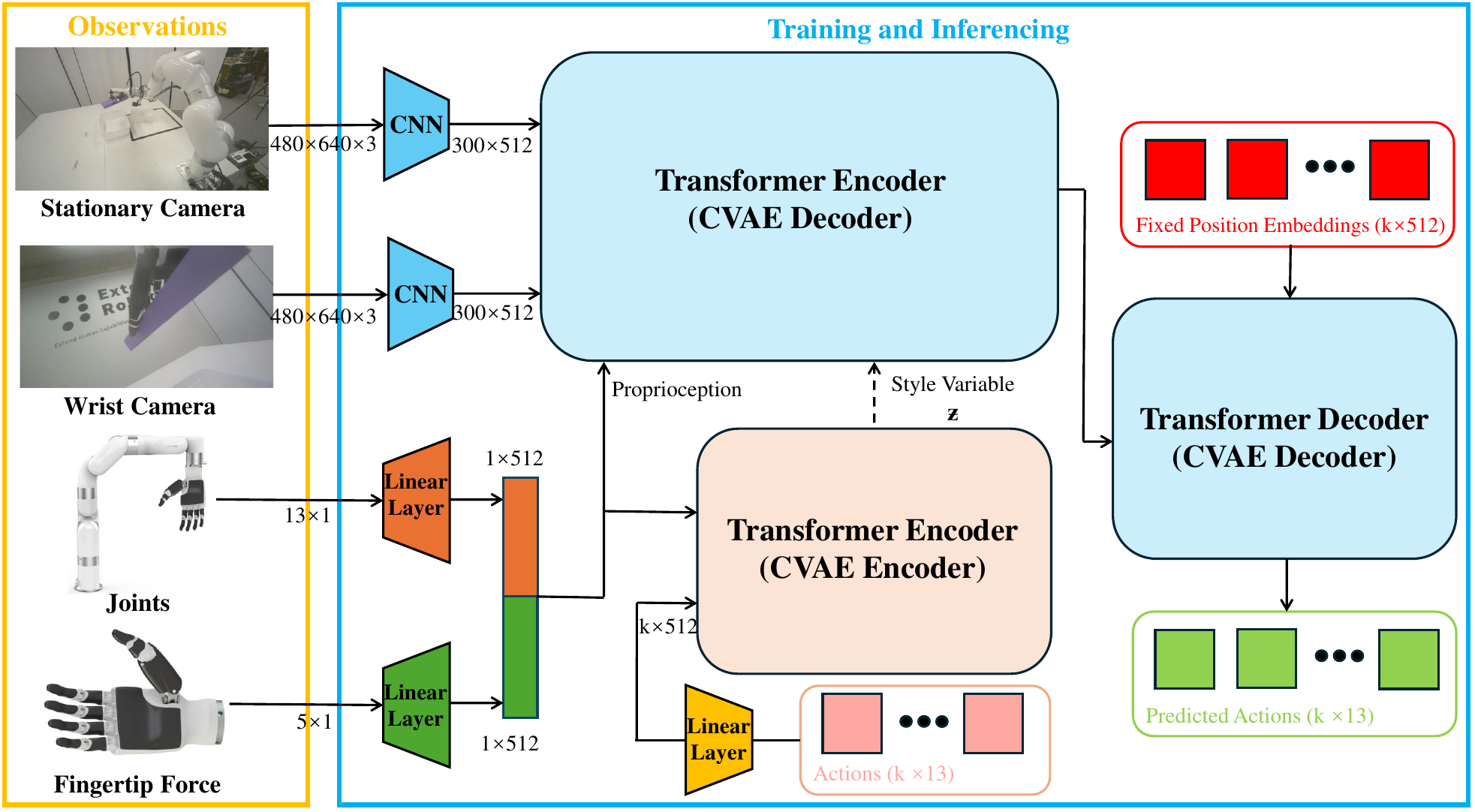}
    \caption{Flowchart of the proposed Haptic-ACT. The observations include RGB images from two cameras, the robot's joint positions, and the fingertip forces of the hand. Note that the transformer encoder (CVAE encoder) operates only during the training phase to compute the style variable for the transformer encoder (CVAE decoder). During the inference phase, the style variable is fixed at 0.}
    \label{fig:2}
    \vspace{-4mm}
\end{figure*}

To address this issue, this work introduces an immersive VR-based setup for teleoperation to collect demonstrations from human demonstrators. Additionally, an imitation learning framework called Haptic-ACT is proposed. The summary diagram of the proposed immersive VR-based setup is shown in Fig.~\ref{fig:1}. The robot system consists of an xArm7 robot arm equipped with an Inspire robot hand and two ZED cameras. The human demonstrator remotely teleoperates the robot using a Meta Quest 3 headset and a SenseGlove, which tracks the hand's movements and maps them to the robot's joint positions using inverse kinematics (IK). The camera feed is rendered in the VR headset, enabling the demonstrator to see the robot's perspective in real-time. The fingertip contact force from the robot is mapped to the SenseGlove motor torque, allowing the demonstrator to experience an immersive demonstration~\cite{Li2023ImmersiveLearning}. The framework of the proposed Haptic-ACT is shown in Fig.~\ref{fig:2}. The observations include RGB images from the cameras, the robot's joint positions, and fingertip contact forces. The haptic information enables the robot to learn how to make soft contact with objects.
    
We summarize our main contributions as follows: (1) A VR-based setup that allows human demonstrators to teleoperate robots immersively. (2) The integration of SenseGlove, which provides haptic feedback to enhance teleoperation. (3) The proposed Haptic-ACT, which enables robots to learn more compliant manipulations compared to the original ACT.

\section{Related Work}\label{sec:related_work}
% - 1. Robot Manipulation
% - 2. Learning-based Robot Manipulation
% - 3. Demonstration Acquisition
\subsection{Robotic Manipulation}\label{sec:manipulation}
Robotic manipulation, encompassing tasks such as grasping, moving, and reorienting objects, is a fundamental capability in robotics. These tasks often require varying levels of contact with the environment, making precise control of contact forces—whether implicitly or explicitly—crucial for successful execution. As robots increasingly take on roles traditionally performed by humans, research on robotic manipulation has expanded significantly~\cite{Song2019AManipulation,Suomalainen2022AContact}.

Manipulation tasks frequently involve contact-rich interactions, such as grasping a hammer for hammering~\cite{Qin2020KETO:Manipulation}, screwing on a bottle cap~\cite{Johns2021Coarse-to-FineDemonstration}, or folding clothes~\cite{Shridhar2021CLIPORT:Manipulation,Chen2024TraKDis:Manipulation}. The most intuitive approach to robotic manipulation is to design controllers based on control theory. Among various control strategies, impedance control is particularly notable for enabling desired dynamic interactions between a manipulator and its environment. This method regulates the dynamic relationship between the manipulator’s motion variables and the contact forces, making it widely used for force tracking~\cite{Roveda2016OptimalTasks}, human-robot interaction~\cite{Li2017AdaptiveSignals}, and other applications. However, traditional controllers often struggle with adaptability, handling only a limited range of tasks and failing in unforeseen situations.

In recent years, learning-based approaches have gained prominence in robotic manipulation~\cite{Yu2022SE-ResUNet:Method,Ni2020PointNet++Clouds,Li2022EfficientGrasp:Hands}. Among these, imitation learning has demonstrated the greatest potential for improving manipulation performance~\cite{Xie2024DecomposingManipulation,Zhang2018DeepTeleoperation}. A key advantage of imitation learning is its ability to leverage human expertise to teach robots complex manipulation skills without requiring explicit programming. Techniques such as behavior cloning and inverse reinforcement learning enable robots to generalize from expert demonstrations and adapt to various tasks~\cite{Rahmatizadeh2018Vision-BasedDemonstration,Argall2009ADemonstration}. However, effectively acquiring high-quality demonstrations remains a significant challenge.

\subsection{Platforms for Demonstrations}\label{sec:platform}
As discussed earlier, effectively gathering demonstrations with an appropriate platform is crucial for acquiring high-quality demonstrations. Recently, the ALOHA platform~\cite{Zhao2023LearningHardware} was designed to provide an affordable and accessible platform for bimanual teleoperation, allowing users to control two robotic arms simultaneously. The newer version of ALOHA includes a mobile platform, allowing it to perform manipulations over a larger area~\cite{Zipeng2024MobileREAL-WORLD}. However, both platforms require the human to be physically present next to the robot, to provide control input, due to lack of effective remote visual perception. \cite{Kim2023TrainingTransfer} proposed a master-to-robot policy transfer system that does not require robots for teaching force feedback-based manipulation tasks. However, the two-finger parallel gripper used in the platform lacks dexterity, limiting its ability to perform certain complex manipulations. To facilitate the demonstration of dexterous manipulations, DexCap was specifically designed for capturing and analyzing tasks involving intricate manipulation~\cite{Wang2024DexCap:Manipulation}. However, DexCap still requires the physical presence of a human demonstrator. To facilitate robotic teleoperation, Open-TeleVision was designed to enhance remote control through immersive and active visual feedback~\cite{Xuxin2024Open-TeleVision:Feedback}. Despite this, Open-TeleVision still lacks haptic feedback,  and suffers from motion sickness due to tightly coupled motion latency. In this work, a VR-based teleoperation platform AMAS, developed by Extend Robotics, is proposed to be integrated with SenseGlove to address the limitations of the prior art. Similar to the approach in~\cite{Li2023ImmersiveLearning}, the incorporation of haptic feedback is expected to enhance the human demonstration experience. Additionally, a new multi-modal architecture is proposed to incorporate vision-based ACT with haptic information with increased dexterity.

\section{Methods}\label{sec:methods}
\subsection{VR-based Demonstration System}\label{sec:VR}
\begin{figure}[t!]
    \centering
    \includegraphics[width=\columnwidth]{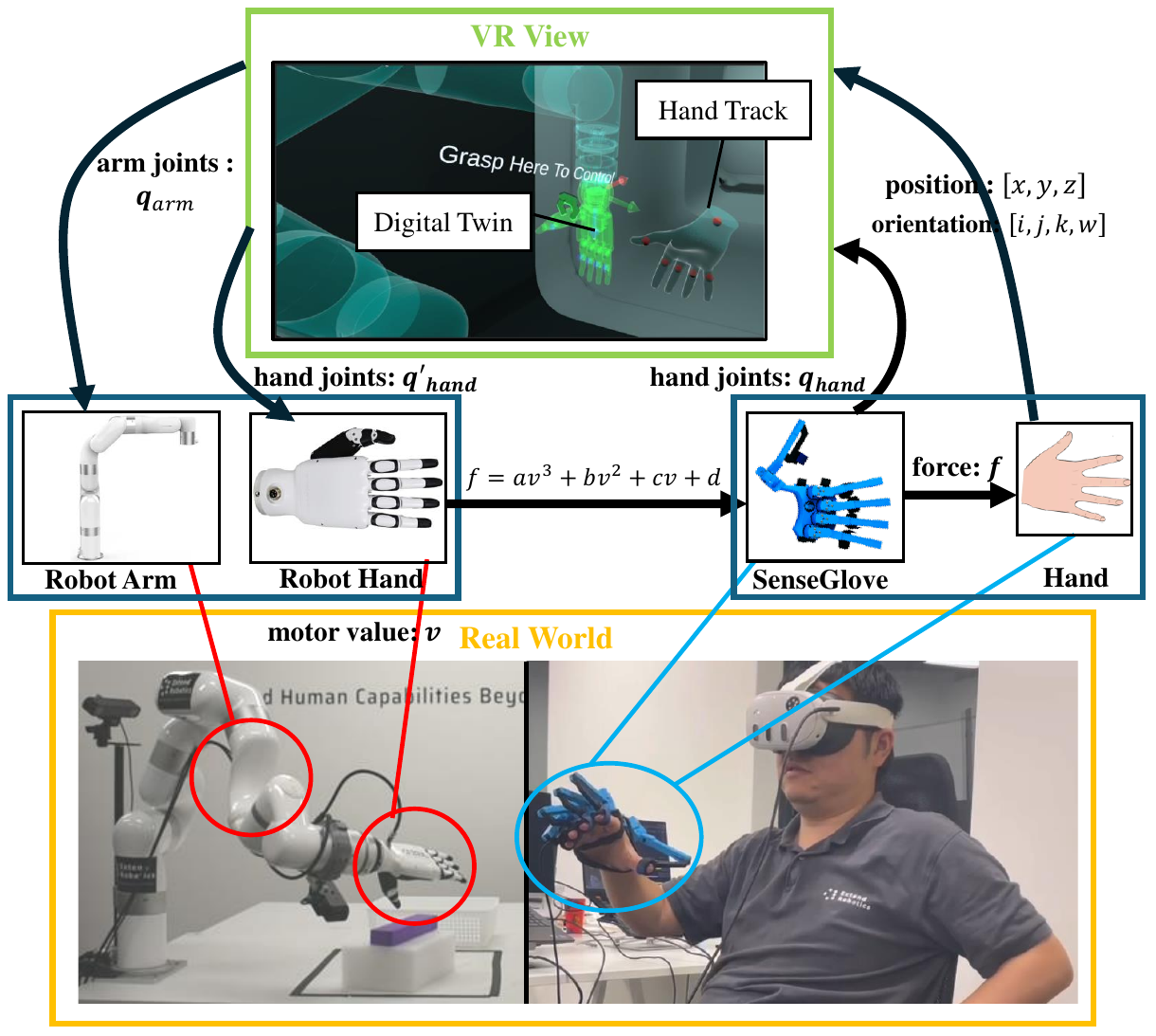}
    \caption{Communication and feedback within the immersive VR-based teleoperation system involve capturing the user's hand position and orientation with the Meta Quest 3. A digital twin is employed to calculate inverse kinematics for the real robot arm. Finger joint positions are captured using a SenseGlove and mapped to desired positions for the real robot hand. All commands are published through ROS, and motor values from the real robot hand are translated into fingertip forces, which are then applied to the user via the SenseGlove.}
    \label{fig:3}
    \vspace{-4mm}
\end{figure}

As mentioned in Section~\ref{sec:platform}, effectively collecting demonstrations with the right platform is crucial for obtaining high-quality results. In this work, we introduce an immersive VR-based platform for gathering demonstrations from human users. As shown in Fig.~\ref{fig:3}, the human demonstrator uses a Meta Quest 3 VR headset and a SenseGlove to remotely control a real robot system. The headset tracks the position $[x,y,z]$ and orientation $[i,j,k,w]$ of the demonstrator's hand, while the SenseGlove captures the movements of the hand joints $\mathbf{q}_{hand}$. This data is then processed in Unity, where we designed a digital twin in Unity to compute the IK for the real robot by solving the following equation:
\begin{equation}
\label{eq:IK}
\mathbf{T}_{base}^{end} \mathbf{\theta} = \mathbf{T}_{target}(\mathbf{p},\mathbf{R})
\end{equation}
where $\mathbf{T}_{base}^{end}$ denotes the transformation matrix from the robot's base frame to the end-effector frame, $\theta$ represents the joint angles of the robot arm, and $ \mathbf{T}_{target}(\mathbf{p},\mathbf{R})$ is the transformation matrix for the target location. The computed joint angles $\mathbf{q}_{arm}$ are sent to the robot arm via ROS. Additionally, the human hand joints $\mathbf{q}_{hand}$ mapped to the robot hand joints $\mathbf{q}'_{hand}$ and also transmitted to the robot hand via ROS. 

To render the robot's view in the VR headset, we use two ZED stereo cameras: one stationary and one mounted on the robot’s wrist. These cameras capture the scene from the robot’s perspective and render it in the VR headset. The advantage of the proposed VR-based platform is that it extends the demonstrator's capabilities beyond physical presence, offering an immersive experience that enhances their ability to interact with the robotic system.

\subsection{Haptic Feedback}\label{sec:haptic}
Although the VR headset allows the demonstrator to teleoperate the robot in a visually immersive way, it still lacks additional sensory feedback that could enhance the overall experience. To address this, we integrated a SenseGlove into the VR setup, enabling the human demonstrator to receive haptic feedback during demonstrations. As shown in Fig.~\ref{fig:3}, the fingertip forces of the real robot hand can be inferred from its motor values. To map these motor values to the corresponding fingertip forces, we used a force gauge to touch each fingertip and recorded the force-motor value pairs for each finger. We then applied data regression methods to derive a formula that converts motor values into corresponding forces, as demonstrated in the following equation:
\begin{align}
\label{eq:torque2force}
    f_i& = a_iv^3+b_iv^2+c_iv+d_i,
\end{align}
where $i$ denotes the number of the finger, $f_i$ and $v_i$ are the fingertip force and motor value of a specific finger, respectively, and $a_i, b_i, c_i, d_i$ are the parameters for the regression formula. The exact values of these parameters are shown in Table.~\ref{table:1}.
\begin{table}[ht]
\centering
\caption{Coefficients for mapping motor values to fingertip forces.}
\begin{tabular}{|c|c|c|c|c|c|}
\hline
\textbf{} & \textbf{Thumb} & \textbf{Index} & \textbf{Middle} & \textbf{Ring} & \textbf{Pinky} \\ \hline
$a$         & 2.25e-9        & 3.23e-10       & 5.51e-10        & -4.98e-10     & 0              \\ \hline
$b$         & -5.28e-6       & -4.18e-7       & -1.88e-6        & 2.40e-6       & 5.73e-7        \\ \hline
$c$         & 8.03e-3        & 2.05e-3        & 3.45e-4         & 1.71e-3       & 1.43e-3        \\ \hline
$d$         & 3.23e-1        & -2.11e-2       & -3.76e-2        & -1.13e-2      & 2.39e-2        \\ \hline
\end{tabular}
\label{table:1}
\end{table}

Similar to~\cite{Li2023ImmersiveLearning}, the fingertip forces are converted into a pulse-width modulation (PWM) signal for the SenseGlove. The duty cycle of this signal is determined by empirically fitting a quadratic curve to the measured force outputs from each resistive tendon:
\begin{equation}
\label{eq:pwm}
    \text{\% duty cycle }i = \sqrt{(f_i - m) / n}
\end{equation}
where $m = 1.72\times10^{-3}$, and $n = 2.57$.

\subsection{Lantency Handling}\label{sec:latency}
In order to handle the potential latency of the system, we apply the following technique to deal with this issue:
\subsubsection{\bf{3D model update and VR rendering decoupling}}
To mitigate latency, the 3D model update is decoupled from VR rendering, preventing motion sickness. For example, the model updates at 30 Hz, while VR rendering runs at 120 Hz, adjusting for head movements. If there’s network lag, the model may be delayed, but this doesn’t cause motion sickness as the rendering remains smooth.
\subsubsection{\bf{Position control based on inverse kinematics}}
The robot is controlled through a digital twin simulated with inverse kinematics in the VR environment. The input for the physical robot is based on the joint poses of the digital twin, updated in real time. The robot controller independently follows the requested state, making it resistant to latency.
\subsubsection{\bf{Behavior in case of increased latency and network problems}}
If latency or network issues occur, the robot follows the state requested by the VR user via the digital twin. While high-bandwidth visual feedback may lag, the user sees both the transparent digital twin (set-point) and the lagging feedback. The misalignment reflects robot inertia and visual feedback latency. Once the visual feedback catches up, the user can be confident the robot has achieved the requested pose, ensuring control despite latency.

\subsection{Haptic-ACT}\label{sec:haptic-act}
Imitation learning methods have been widely used for enabling robots to learn manipulation tasks from demonstrations. Recently, Action Chunking with Transformers (ACT)~\cite{Zhao2023LearningHardware} was introduced, offering efficient handling of long-horizon tasks by segmenting them into smaller, manageable chunks, thereby improving task performance and learning efficiency in robotic manipulation. However, ACT lacks integration of haptic information, which is crucial for contact-rich manipulations. To address this, we propose the Haptic-ACT framework to incorporate haptic feedback for enhanced learning and performance. As shown in Fig.~\ref{fig:2}, Haptic-ACT takes two $480\times640\times3$ RGB images from two cameras, $13\times1$ joint positions (7 for arm and 6 for hand), and $5\times1$ fingertip forces as observations. Before the observations are fed into the networks, all the data are normalized using min-max normalization as follows:
\begin{equation}
\label{eq:normalization}
o_{\text{norm}} = \frac{o - o_{\text{mean}}}{o_{\text{std}}}
\end{equation}
where $o$ represents the original data, $o_{\text{mean}}$ and $o_{\text{std}}$ are the mean and deviation values of the data, respectively, and $o_{\text{norm}}$ is the normalized value. The normalized observations are embedded using Convolutional Neural Networks (CNNs) or linear layers and are then fed into the transformer encoders. During training, the Conditional Variational AutoEncoder (CVAE) encoder generates a style variable for the CVAE decoder. During inference, the style variable is set to zero for deterministic decoding. The action sequence is represented as $k\times13$, where $k$ refers to the manually defined chunk size. The transformers are optimized by minimizing the Mean Squared Error (MSE) between the predicted actions and the ground truth actions, as well as the Kullback-Leibler (KL) Divergence between the encoder output and a standard normal distribution. The loss function can be summarized as follows:
\begin{equation}
\label{eq:loss}
\mathcal{L} = \text{MSE}\left(\hat{a}_i, a_i  \right) + \beta\text{KL}\left(q(z \mid a_i,\overline{o}_{\text{norm}}) \,\| \,p(z)\right),
\end{equation}
where $\hat{a}_i$ and $a_i$ represent the $i$-th chunk of predicted actions and ground truth actions, respectively; $q(z \mid a_i, \overline{o}_t)$ denotes the encoder distribution; $p(z)$ is the standard normal distribution; and $\beta$ is a weighting coefficient for the KL divergence term.

\section{Experimental Setup}\label{sec:experimental_setup}

\begin{figure}[t!]
    \vspace{4mm}
    \centering
    \includegraphics[width=0.45\textwidth]{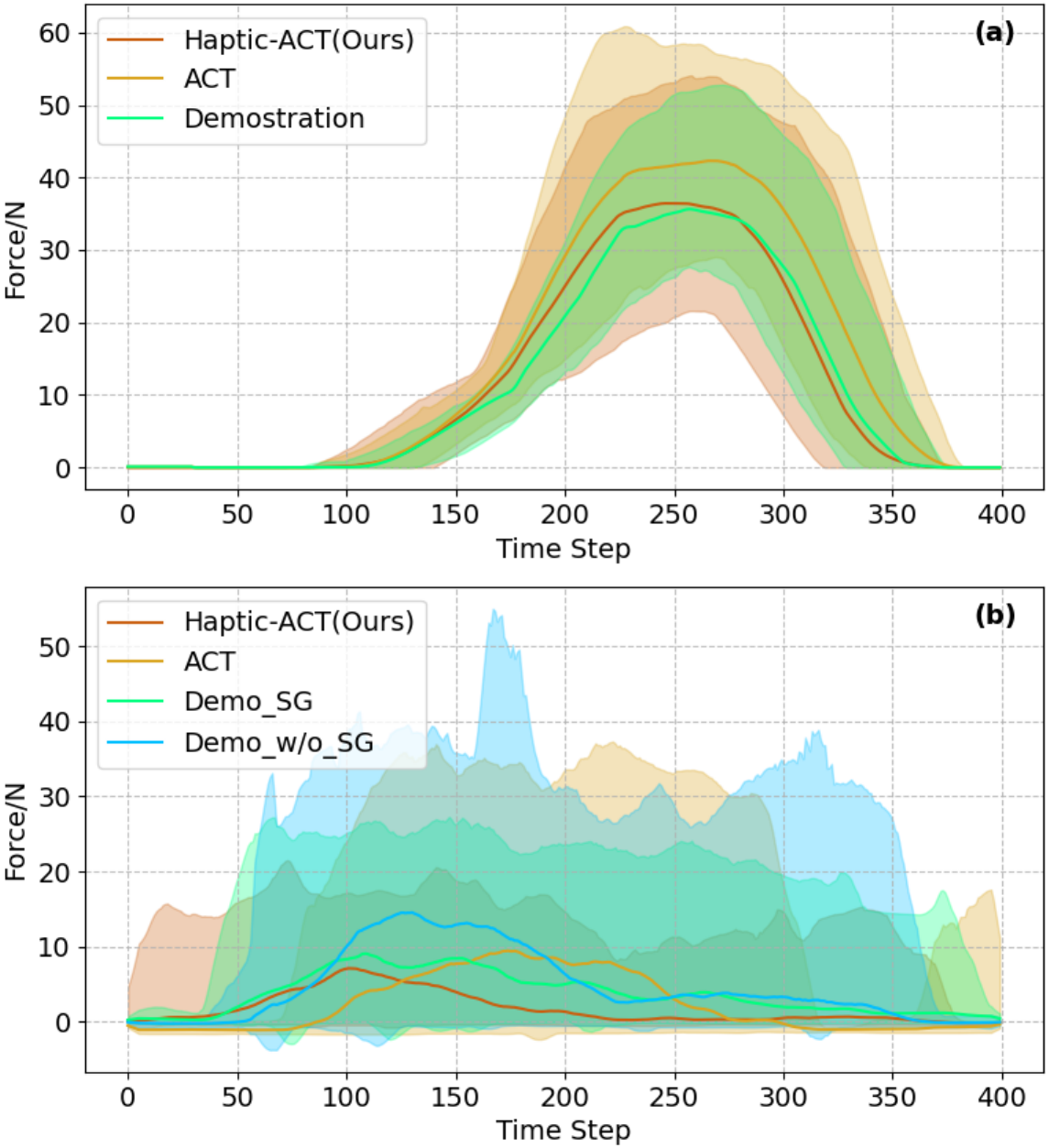}
    \caption{Average fingertip force during manipulation. \textbf{(a)} Displays the results from the MuJoCo simulator. \textbf{(b)} Presents the results from the real-world experiment, where Demo\_SG indicates demonstrations with SenseGlove, and Demo\_w\slash o\_SG refers to demonstrations without SenseGlove.}
    \label{fig:4}
    \vspace{-4mm}
\end{figure}

\begin{figure}[t!]
    \vspace{4mm}
    \centering
    \includegraphics[width=0.45\textwidth]{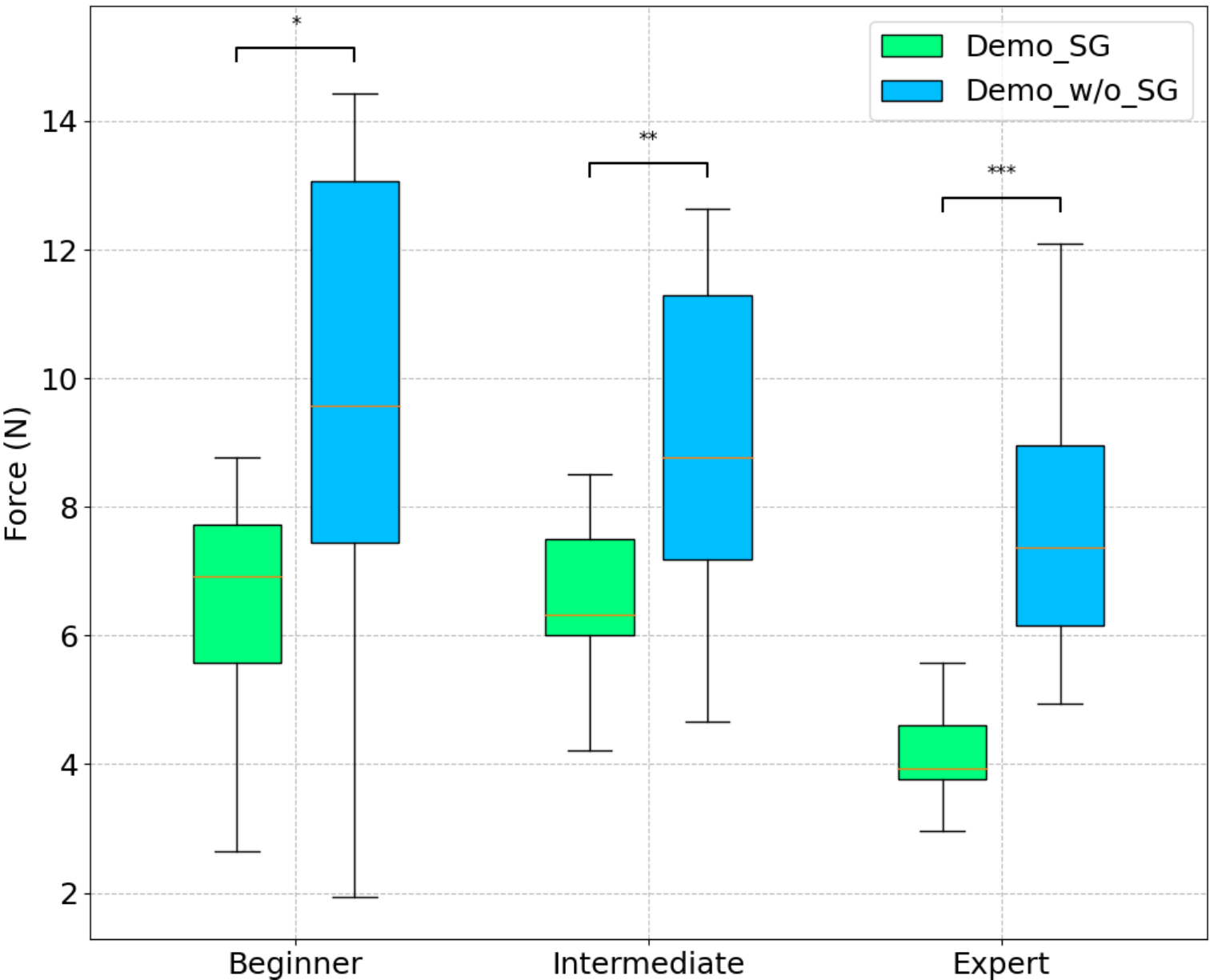}
    \caption{Comparison of average fingertip force among different demonstrator groups during a real-world pick-and-place task. The results of Student’s $t$-test are indicated: ***$p < 0.001$, **$p < 0.01$, *$p<0.05$.}
    \label{fig:5}
    \vspace{-4mm}
\end{figure}

\begin{figure*}
    \vspace{4mm}
    \centering
    \includegraphics[width=0.8\textwidth]{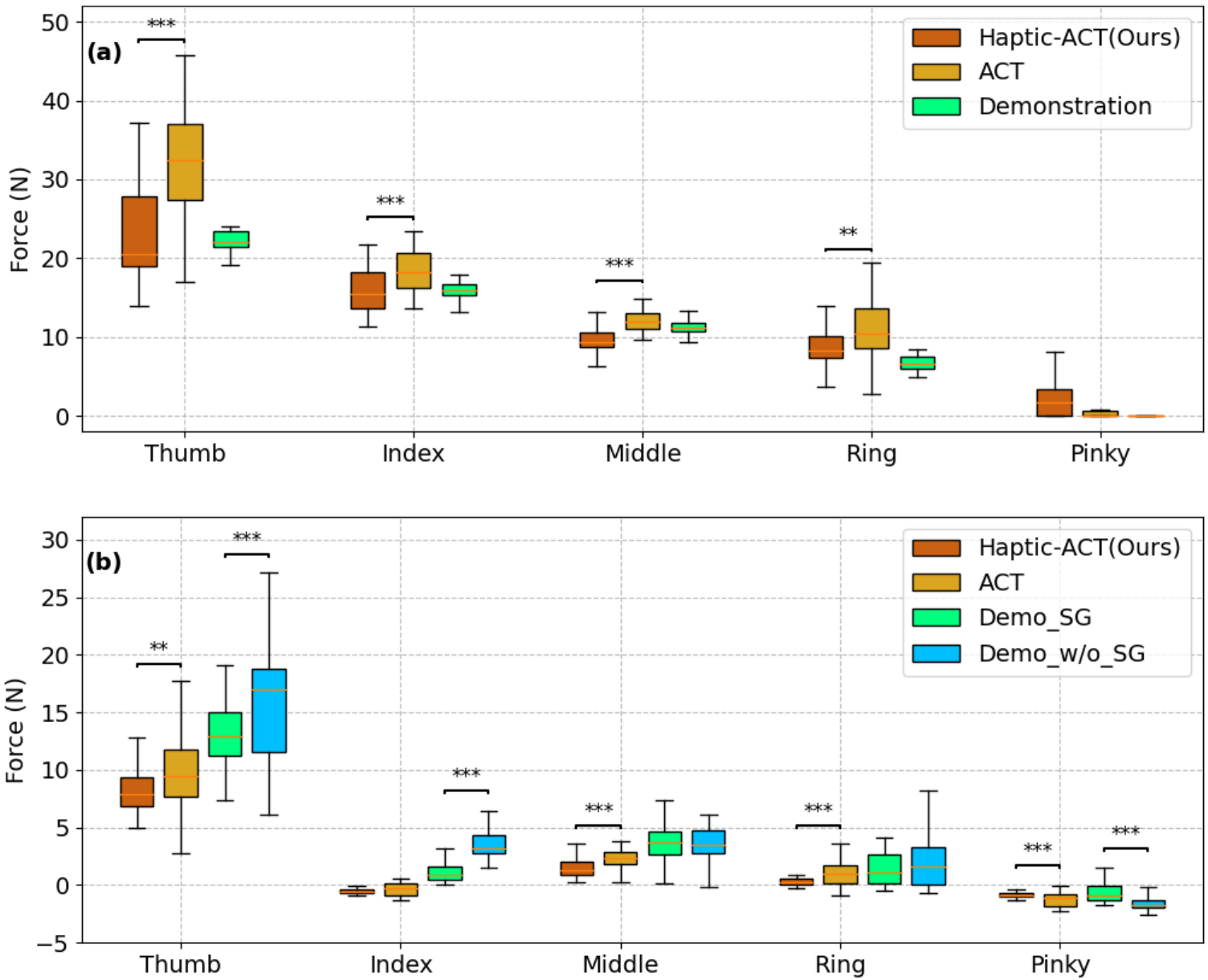}
    \caption{Box plots of fingertip forces for each finger during manipulations, comparing the performances of the proposed Haptic-ACT, the original ACT, and demonstrations. \textbf{(a)} Shows results from the MuJoCo simulator. \textbf{(b)} Shows results from real-world experiments, where Demo\_SG denotes demonstrations with SenseGlove and Demo\_w\slash o\_SG denotes demonstrations without SenseGlove. Results of student's $t$-test, ***$p < 0.001$, **$p < 0.01$, *$p<0.05$.}
    \label{fig:6}
    \vspace{-4mm}
\end{figure*}

We evaluated the VR-based platform and Haptic-ACT framework in both MuJoCo simulation~\cite{Emanuel2012MuJoCo:Control} and real-world experiments. The simulation featured an Xarm7 with an Inspire Robots dexterous hand, allowing controlled testing before real-world validation. For physical experiments, we used the VR-based teleoperation system from Section~\ref{sec:methods}.

In the MuJoCo simulator, we evaluated Haptic-ACT on a simple pick-and-place task, where the robot picks up a block and places it into a basket. A scripted policy was used to perform 50 successful episodes for training, each lasting 400 timesteps, with the simulation running at 50 FPS. The observations collected included an RGB image from a front-facing camera, the robot's joint positions, and fingertip forces. Using these collected demonstrations, we trained policies with ACT and Haptic-ACT, respectively. We then evaluated the fingertip forces generated by the trained policies and compared the results among ACT, Haptic-ACT, and the original demonstrations. The training was conducted using an NVIDIA GeForce RTX3070ti GPU.

For the real-world experiment, we utilized the VR-based platform to conduct the same pick-and-place task with human demonstrators. Demonstrations were collected under two haptic conditions: with SenseGlove and without SenseGlove, to assess the impact of haptic feedback. As in the simulation, we gathered 50 successful episodes for each haptic condition, with each episode spanning 400 timesteps and data recorded at a frequency of 15Hz. To enhance the observations, we included an additional wrist camera image. Policies based on both the ACT and Haptic-ACT frameworks were trained and tested on the robot setup to evaluate the effectiveness, using an NVIDIA GeForce RTX4070 GPU for training.

For both ACT and Haptic-ACT training, the learning rate is set to 0.00001, with a chunk size of 50 and a batch size of 8. The backbone network is ResNet18, while the encoder and decoder consist of 4 and 7 layers, respectively. The models are trained for 2000 epochs to ensure convergence and optimal performance.

\begin{table}[ht]
\centering
\caption{Success rate of a pick-and-place task.}
\begin{tabular}{|c|c|c|c|}
\hline
                & Attempts & Success & Success rate (\%) \\ \hline
ACT\_sim        & 55       & 50      & 90.9              \\ \hline
Haptic-ACT\_sim & 55       & 50      & 90.9              \\ \hline
ACT\_rw         & 60       & 50      & 83.3              \\ \hline
Haptic-ACT\_rw  & 58       & 50      & 86.2              \\ \hline
\end{tabular}
\label{table:2}
\end{table}

\section{Results and Discussion}\label{sec:results_discussion}
\subsection{Simulation Results}
We first evaluated the performance of Haptic-ACT and ACT in the MuJoCo simulator. As shown in Fig.~\ref{fig:4}\textbf{(a)}, the average contact forces for the five fingers during 50 pick-and-place manipulations were analyzed. The results indicate that both the average and distributed forces for Haptic-ACT are approximately 15\% lower than those for ACT, suggesting that Haptic-ACT achieves more compliant and softer grasps compared to the original ACT. Additionally, Haptic-ACT more closely mimics the demonstrations, as indicated by the similarity between the contact force curves and those observed in the demonstrations. Table~\ref{table:2} indicates that despite contact force is reduced significantly, the success rate remains the same for Haptic-ACT and ACT.

A Student’s t-test was also conducted to compare the contact forces between Haptic-ACT and ACT. As shown in Fig.~\ref{fig:6}\textbf{(a)}, the thumb in Haptic-ACT exhibits significantly lower (approximately 30\% lower) contact force than in ACT (p \textless \ 0.001), with similar reductions observed for the index and middle fingers (p \textless \ 0.01). The ring finger also shows a notable decrease (p \textless \ 0.01), while the pinky remains statistically unchanged. However, the pinky force remains relatively close to zero and does not significantly affect the grasping performance. Thus, the slightly elevated pinky force is acceptable and does not detract from the overall effectiveness of the grasp. These results confirm that Haptic-ACT enables softer and more compliant grasps, closely resembling human demonstrations and improving manipulation of delicate objects.

In summary, the simulation results demonstrate that Haptic-ACT achieves softer grasps in the pick-and-place manipulation task, which is particularly beneficial for handling deformable objects such as paper cups and fruits, where gentle and controlled grasping is essential to avoid damage and ensure effective manipulation. This improvement suggests that Haptic-ACT could be highly beneficial in real-world applications requiring delicate handling.

\subsection{Data Collection Results}
We collected 50 episodes of demonstrations using the proposed immersive VR-based platform, both with and without SenseGlove. Five individuals were selected to provide demonstration data, and they were categorized based on their proficiency with the experimental equipment into three groups: beginner, intermediate, and expert. The groups consisted of one beginner, two intermediate participants, and two expert participants. Each individual completed 10 episodes of the pick-and-place task for each haptic condition, during which we recorded the contact force, robot joint states, and two RGB images from a stationary camera and a wrist camera, respectively. Our primary focus was to investigate the fingertip force exerted during task execution. Fig.~\ref{fig:5} presents the average fingertip force applied by each group, offering a visual representation of the differences in force across proficiency levels. "Demo\_SG" and "Demo\_w\slash o\_SG" represent the results from demonstrations with and without SenseGlove, respectively.

It is evident from Fig.~\ref{fig:5} that all groups of demonstrators tend to apply a larger fingertip force when picking up the object, which could potentially damage more delicate objects. However, with the haptic feedback provided by SenseGlove, demonstrators are less likely to apply excessive force, leading to a 25\% reduction in contact force compared to demonstrations without SenseGlove. Additionally, the figure shows that as demonstrators become more familiar with the platform, they tend to apply less contact force, demonstrating more control. Moreover, experienced demonstrators apply force with less variation, indicating increased consistency in their actions. This suggests that with practice, demonstrators not only optimize the force applied but also improve the overall quality of their movements. The feedback from SenseGlove likely accelerates this learning process by providing real-time haptic cues that enhance precision.

The results of the Student’s t-test revealed statistically significant differences, as indicated by the p-values in Fig.\ref{fig:5}: ***$p < 0.001$, **$p < 0.01$, and *$p < 0.05$. These p-values suggest that the observed differences in fingertip force between the groups are unlikely to have occurred by chance. Specifically, the highly significant p-values ($p < 0.001$) observed in comparisons involving the expert group indicate a strong distinction in force application between experts and the other groups, likely reflecting their greater control and precision in handling the teleoperation equipment. The p-values of $p < 0.01$ and $p < 0.05$ in other comparisons further support the notion that proficiency levels influence force application, with intermediate participants showing moderate differences compared to beginners and experts.

\subsection{Real-World Results}
\begin{figure*}[t]
    \vspace{4mm}
    \centering
    \includegraphics[width=\textwidth]{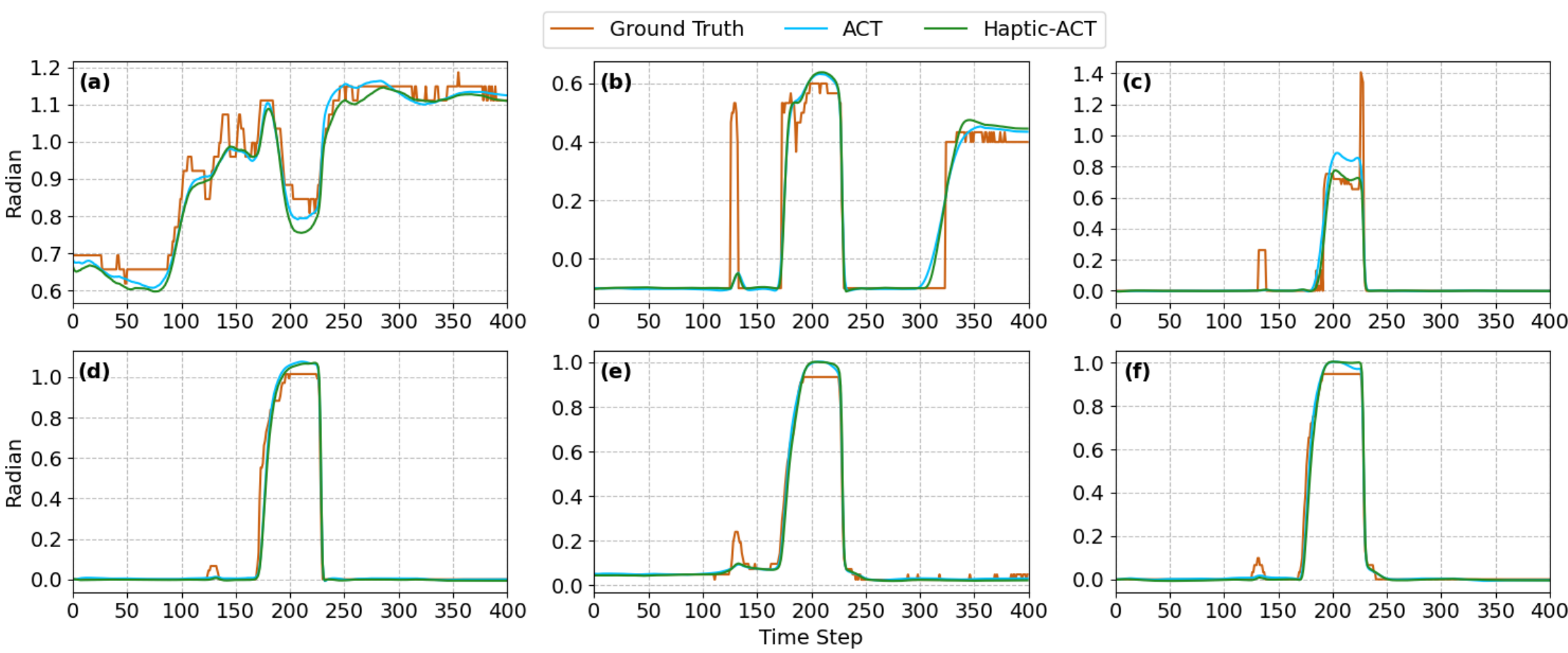}
    \caption{Inferencing results for the hand joint positions are presented, comparing Haptic-ACT, the original ACT, and the ground truth from the demonstrations. Subplots \textbf{(a)} through \textbf{(f)} illustrate the joint positions for the thumb yaw, thumb pitch, index, middle, ring, and pinky fingers, respectively.}
    \label{fig:7}
    \vspace{-4mm}
\end{figure*}

The contact forces under these two haptic conditions are further evaluated in Fig.\ref{fig:4}\textbf{(b)} and Fig.\ref{fig:6}\textbf{(b)}, where "Demo\_SG" and "Demo\_w\slash o\_SG" represent the results from demonstrations with and without SenseGlove, respectively. The plots clearly show that the demonstrators tend to apply less force when haptic feedback is provided. Additionally, the box plots indicate that the distribution of contact forces is narrower with haptic feedback compared to without, suggesting a more consistent and controlled grasping behavior. This narrower distribution is beneficial for handling delicate objects and improving overall manipulation precision. The results of the Student’s t-test further support this observation, indicating that the reduction in fingertip force is unlikely to have occurred by chance, as the p-values for the thumb, index, and pinky fingers are all smaller than 0.001.

The Haptic-ACT and ACT frameworks were also evaluated on the real robot setup. As shown in Fig.\ref{fig:4}\textbf{(b)}, the average contact forces exerted by Haptic-ACT across the five fingers are approximately 15\% lower than those produced by ACT, consistent with the results observed in the simulation. The contact forces on each finger during manipulation were further analyzed in Fig.\ref{fig:6}\textbf{(b)}. Similar to the simulation results, the thumb applied the largest contact force, while the pinky played a minimal role. The results of the Student’s t-test support this observation, showing that the reduction in fingertip force is unlikely to have occurred by chance, as the p-values for all fingers except the index are smaller than 0.01 or 0.001. The lower contact forces achieved by Haptic-ACT not only demonstrate its ability to produce softer grasps but also suggest that it may be more suitable for handling delicate or fragile objects in real-world applications, reducing the risk of damage during manipulation tasks. Table~\ref{table:2} shows that, despite the significant reduction in contact force, the success rate remains similar for both Haptic-ACT and ACT.

Finally, we investigated the potential reasons behind Haptic-ACT's superior performance compared to ACT, as shown in Fig.~\ref{fig:7}. This figure presents an inferencing result for the hand joint positions of Haptic-ACT, ACT, and the ground truth from the demonstrations. Subplots \textbf{(a)} through \textbf{(f)} depict the joint positions for the thumb yaw, thumb pitch, index, middle, ring, and pinky fingers, respectively. It is evident that the most significant variations between ACT and Haptic-ACT occur in the thumb yaw and index finger joints. In particular, the joint positions in Haptic-ACT tend to be slightly smaller than those in ACT, leading to less applied contact force on the object during manipulation. 

\section{Conclusions and Future Work}\label{sec:conclusions}
In this work, we introduced an immersive VR-based teleoperation setup designed to collect demonstrations from human users. By comparing the fingertip forces during demonstrations using the proposed platform with and without SenseGlove, we concluded that haptic feedback enables the demonstrator to perform tasks with less effort and achieve more precise manipulations. Furthermore, we proposed an imitation learning framework called Haptic-ACT, which leverages haptic feedback to improve manipulation performance. Through extensive experiments conducted in both a simulated environment and on a real robot setup, we demonstrated that Haptic-ACT achieves compliant grasps that indicates better imitating human demonstrations compared to the original ACT framework. This improvement is particularly important for tasks involving delicate or deformable objects, where precise force control is essential to avoid damage.

In future work, we plan to evaluate the VR platform and Haptic-ACT on more complex tasks such as drilling, brushing, and pouring water. Additionally, we aim to integrate 3D visual data and language models to further enhance Haptic-ACT's performance and adaptability across a broader range of tasks, enabling it to handle more delicate and intricate manipulations with greater precision.

\section*{Acknowledgement}
The authors would like to thank the Dyson School of Design Engineering, Imperial College London for the technical support of this work, and also thank Extend Robotics for providing the equipment used in this work.

\newpage
\addtolength{\textheight}{-5.0cm}  % This command serves to balance the column lengths
                                  % on the last page of the document manually. It shortens
                                  % the textheight of the last page by a suitable amount.
                                  % This command does not take effect until the next page
                                  % so it should come on the page before the last. Make
                                  % sure that you do not shorten the textheight too much.

%%%%%%%%%%%%%%%%%%%%%%%%%%%%%%%%%%%%%%%%%%%%%%%%%%%%%%%%%%%%%%%%%%%%%%%%%%%%%%%%

\bibliographystyle{IEEEtran}
\bibliography{references}

\begin{thebibliography}{10}
\providecommand{\url}[1]{#1}
\csname url@rmstyle\endcsname
\providecommand{\newblock}{\relax}
\providecommand{\bibinfo}[2]{#2}
\providecommand\BIBentrySTDinterwordspacing{\spaceskip=0pt\relax}
\providecommand\BIBentryALTinterwordstretchfactor{4}
\providecommand\BIBentryALTinterwordspacing{\spaceskip=\fontdimen2\font plus
\BIBentryALTinterwordstretchfactor\fontdimen3\font minus \fontdimen4\font\relax}
\providecommand\BIBforeignlanguage[2]{{%
\expandafter\ifx\csname l@#1\endcsname\relax
\typeout{** WARNING: IEEEtran.bst: No hyphenation pattern has been}%
\typeout{** loaded for the language `#1'. Using the pattern for}%
\typeout{** the default language instead.}%
\else
\language=\csname l@#1\endcsname
\fi
#2}}

\bibitem{Cui2021TowardManipulation}
J.~Cui and J.~Trinkle, ``{Toward next-generation learned robot manipulation},'' \emph{Sci. Robot}, vol.~6, p. 9461, 2021.

\bibitem{Shridhar2021CLIPORT:Manipulation}
M.~Shridhar, L.~Manuelli, and D.~Fox, ``{CLIPORT: What and Where Pathways for Robotic Manipulation},'' in \emph{Conference on Robot Learning (CoRL)}, 2021.

\bibitem{Johns2021Coarse-to-FineDemonstration}
E.~Johns, ``{Coarse-to-Fine Imitation Learning: Robot Manipulation from a Single Demonstration},'' in \emph{2021 IEEE International Conference on Robotics and Automation (ICRA)}.\hskip 1em plus 0.5em minus 0.4em\relax IEEE, 2021, pp. 4613--4619.

\bibitem{Li2022EfficientGrasp:Hands}
K.~Li, N.~Baron, X.~Zhang, and N.~Rojas, ``{EfficientGrasp: A Unified Data-Efficient Learning to Grasp Method for Multi-Fingered Robot Hands},'' \emph{IEEE Robotics and Automation Letters}, vol.~7, no.~4, pp. 8619--8626, 10 2022.

\bibitem{Wu2022LearningVisualization}
Y.-H. Wu, J.~Wang, and X.~Wang, ``{Learning Generalizable Dexterous Manipulation from Human Grasp Affordance Learned Policy Visualization on Unseen Test Objects Affordance Demonstrations for Training Object Human Grasp Robot Grasp Object Human Grasp Robot Grasp Object Policy Visualization},'' in \emph{Conference on Robot Learning (CoRL)}, 2022.

\bibitem{Kwon2024LanguageGenerators}
T.~Kwon, N.~Di~Palo, and E.~Johns, ``{Language Models as Zero-Shot Trajectory Generators},'' \emph{IEEE Robotics and Automation Letters}, vol.~9, no.~7, pp. 6728--6735, 7 2024.

\bibitem{Jiang2023VIMA:Prompts}
Y.~Jiang, A.~Gupta, Z.~Zhang, G.~Wang, Y.~Dou, Y.~Chen, L.~Fei-Fei, A.~Anandkumar, Y.~Zhu, and L.~Fan, ``{VIMA: General Robot Manipulation with Multimodal Prompts},'' in \emph{International Conference on Machine Learning}, 2023.

\bibitem{Yu2022SE-ResUNet:Method}
S.~Yu, D.-H. Zhai, Y.~Xia, H.~Wu, and J.~Liao, ``{SE-ResUNet: A Novel Robotic Grasp Detection Method},'' \emph{IEEE Robotics and Automation Letters}, vol.~7, no.~2, pp. 5238--5245, 4 2022.

\bibitem{An2024RGBManip:Estimation}
B.~An, Y.~Geng, K.~Chen, X.~Li, Q.~Dou, and H.~Dong, ``{RGBManip: Monocular Image-based Robotic Manipulation through Active Object Pose Estimation},'' in \emph{IEEE International Conference on Robotics and Automation}, 2024.

\bibitem{Ni2020PointNet++Clouds}
P.~Ni, W.~Zhang, X.~Zhu, and Q.~Cao, ``{PointNet++ Grasping: Learning An End-to-end Spatial Grasp Generation Algorithm from Sparse Point Clouds},'' in \emph{2020 IEEE International Conference on Robotics and Automation (ICRA)}.\hskip 1em plus 0.5em minus 0.4em\relax IEEE, 5 2020, pp. 3619--3625.

\bibitem{Liang2019PointNetGPD:Sets}
H.~Liang, X.~Ma, S.~Li, M.~Gorner, S.~Tang, B.~Fang, F.~Sun, and J.~Zhang, ``{PointNetGPD: Detecting Grasp Configurations from Point Sets},'' in \emph{2019 International Conference on Robotics and Automation (ICRA)}.\hskip 1em plus 0.5em minus 0.4em\relax IEEE, 5 2019, pp. 3629--3635.

\bibitem{Zhao2021REGNet:Clouds}
B.~Zhao, H.~Zhang, X.~Lan, H.~Wang, Z.~Tian, and N.~Zheng, ``{REGNet: REgion-based Grasp Network for End-to-end Grasp Detection in Point Clouds},'' in \emph{2021 IEEE International Conference on Robotics and Automation (ICRA)}.\hskip 1em plus 0.5em minus 0.4em\relax IEEE, 5 2021, pp. 13\,474--13\,480.

\bibitem{Shao2020UniGrasp:Hands}
L.~Shao, F.~Ferreira, M.~Jorda, V.~Nambiar, J.~Luo, E.~Solowjow, J.~A. Ojea, O.~Khatib, and J.~Bohg, ``{UniGrasp: Learning a Unified Model to Grasp With Multifingered Robotic Hands},'' \emph{IEEE Robotics and Automation Letters}, vol.~5, no.~2, pp. 2286--2293, 4 2020.

\bibitem{Duan2024Manipulate-Anything:Models}
J.~Duan, W.~Yuan, W.~Pumacay, Y.~R. Wang, K.~Ehsani, D.~Fox, and R.~Krishna, ``{Manipulate-Anything: Automating Real-World Robots using Vision-Language Models},'' in \emph{Conference on Robot Learning (CoRL)}, 2024.

\bibitem{Hua2021LearningLearning}
J.~Hua, L.~Zeng, G.~Li, and Z.~Ju, ``{Learning for a robot: Deep reinforcement learning, imitation learning, transfer learning},'' pp. 1--21, 2 2021.

\bibitem{Kim2021Transformer-basedManipulation}
H.~Kim, Y.~Ohmura, and Y.~Kuniyoshi, ``{Transformer-based deep imitation learning for dual-arm robot manipulation},'' in \emph{IEEE International Conference on Intelligent Robots and Systems}.\hskip 1em plus 0.5em minus 0.4em\relax Institute of Electrical and Electronics Engineers Inc., 2021, pp. 8965--8972.

\bibitem{Belkhale2023HYDRA:Learning}
S.~Belkhale, Y.~Cui, and D.~Sadigh, ``{HYDRA: Hybrid Robot Actions for Imitation Learning},'' in \emph{Conference on Robot Learning (CoRL)}, 2023.

\bibitem{Xie2024DecomposingManipulation}
A.~Xie, L.~Lee, T.~Xiao, and C.~Finn, ``{Decomposing the Generalization Gap in Imitation Learning for Visual Robotic Manipulation},'' in \emph{IEEE International Conference on Robotics and Automation (ICRA)}, 2024.

\bibitem{Li2023ImmersiveLearning}
K.~Li, D.~Chappell, and N.~Rojas, ``{Immersive Demonstrations are the Key to Imitation Learning},'' in \emph{Proceedings - IEEE International Conference on Robotics and Automation}, vol. 2023-May.\hskip 1em plus 0.5em minus 0.4em\relax Institute of Electrical and Electronics Engineers Inc., 2023, pp. 5071--5077.

\bibitem{Song2019AManipulation}
P.~Song, Y.~Yu, and X.~Zhang, ``{A Tutorial Survey and Comparison of Impedance Control on Robotic Manipulation},'' \emph{Robotica}, vol.~37, no.~5, pp. 801--836, 5 2019.

\bibitem{Suomalainen2022AContact}
M.~Suomalainen, Y.~Karayiannidis, and V.~Kyrki, ``{A survey of robot manipulation in contact},'' \emph{Robotics and Autonomous Systems}, vol. 156, 10 2022.

\bibitem{Qin2020KETO:Manipulation}
Z.~Qin, K.~Fang, Y.~Zhu, L.~Fei-Fei, and S.~Savarese, ``{KETO: Learning Keypoint Representations for Tool Manipulation},'' in \emph{2020 IEEE International Conference on Robotics and Automation (ICRA)}.\hskip 1em plus 0.5em minus 0.4em\relax IEEE, 5 2020, pp. 7278--7285.

\bibitem{Chen2024TraKDis:Manipulation}
W.~Chen and N.~Rojas, ``{TraKDis: A Transformer-Based Knowledge Distillation Approach for Visual Reinforcement Learning with Application to Cloth Manipulation},'' \emph{IEEE Robotics and Automation Letters}, vol.~9, no.~3, pp. 2455--2462, 3 2024.

\bibitem{Roveda2016OptimalTasks}
L.~Roveda, N.~Iannacci, F.~Vicentini, N.~Pedrocchi, F.~Braghin, and L.~M. Tosatti, ``{Optimal Impedance Force-Tracking Control Design With Impact Formulation for Interaction Tasks},'' \emph{IEEE Robotics and Automation Letters}, vol.~1, no.~1, pp. 130--136, 1 2016.

\bibitem{Li2017AdaptiveSignals}
Z.~Li, Z.~Huang, W.~He, and C.~Y. Su, ``{Adaptive impedance control for an upper limb robotic exoskeleton using biological signals},'' \emph{IEEE Transactions on Industrial Electronics}, vol.~64, no.~2, pp. 1664--1674, 2 2017.

\bibitem{Zhang2018DeepTeleoperation}
T.~Zhang, Z.~McCarthy, O.~Jow, D.~Lee, X.~Chen, K.~Goldberg, and P.~Abbeel, ``{Deep Imitation Learning for Complex Manipulation Tasks from Virtual Reality Teleoperation},'' in \emph{2018 IEEE International Conference on Robotics and Automation (ICRA)}.\hskip 1em plus 0.5em minus 0.4em\relax IEEE, 5 2018, pp. 5628--5635.

\bibitem{Rahmatizadeh2018Vision-BasedDemonstration}
R.~Rahmatizadeh, P.~Abolghasemi, L.~Boloni, and S.~Levine, ``{Vision-Based Multi-Task Manipulation for Inexpensive Robots Using End-to-End Learning from Demonstration},'' in \emph{2018 IEEE International Conference on Robotics and Automation (ICRA)}.\hskip 1em plus 0.5em minus 0.4em\relax IEEE, 5 2018, pp. 3758--3765.

\bibitem{Argall2009ADemonstration}
B.~D. Argall, S.~Chernova, M.~Veloso, and B.~Browning, ``{A survey of robot learning from demonstration},'' \emph{Robotics and Autonomous Systems}, vol.~57, no.~5, pp. 469--483, 5 2009.

\bibitem{Zhao2023LearningHardware}
T.~Z. Zhao, V.~Kumar, S.~Levine, and C.~Finn, ``{Learning Fine-Grained Bimanual Manipulation with Low-Cost Hardware},'' in \emph{Robotics: Science and Systems}, 2023.

\bibitem{Zipeng2024MobileREAL-WORLD}
F.~Zipeng, Z.~Z. Tony, and F.~Chelsea, ``{Mobile ALOHA: Learning Bimanual Mobile Manipulation using Low-Cost Whole-Body Teleoperation PLEASE CHECK THE SUPPLEMENTARY MATERIAL FOR REAL-WORLD},'' in \emph{Conference on Robot Learning (CoRL)}, 2024.

\bibitem{Kim2023TrainingTransfer}
H.~Kim, Y.~Ohmura, A.~Nagakubo, and Y.~Kuniyoshi, ``{Training Robots Without Robots: Deep Imitation Learning for Master-to-Robot Policy Transfer},'' \emph{IEEE Robotics and Automation Letters}, vol.~8, no.~5, pp. 2906--2913, 5 2023.

\bibitem{Wang2024DexCap:Manipulation}
C.~Wang, H.~Shi, W.~Wang, R.~Zhang, L.~Fei-Fei, and C.~K. Liu, ``{DexCap: Scalable and Portable Mocap Data Collection System for Dexterous Manipulation},'' \emph{arXiv preprint arXiv:2403.07788}, 3 2024.

\bibitem{Xuxin2024Open-TeleVision:Feedback}
C.~Xuxin, L.~Jialong, Y.~Shiqi, Y.~Ge, and W.~Xiaolong, ``{Open-TeleVision: Teleoperation with Immersive Active Visual Feedback},'' \emph{arXiv preprint arXiv:2407.01512}, 2024.

\bibitem{Emanuel2012MuJoCo:Control}
T.~Emanuel, E.~Tom, and T.~Yuval, ``{MuJoCo: A physics engine for model-based control},'' in \emph{2012 IEEE/RSJ International Conference on Intelligent Robots and Systems (IROS)}.\hskip 1em plus 0.5em minus 0.4em\relax IEEE, 2012.

\end{thebibliography}

\end{document}